\documentclass[submission,copyright,creativecommons]{eptcs}
\providecommand{\event}{FLACOS 2011} 
\usepackage{breakurl}             

\usepackage{amssymb}
\usepackage{graphicx}
\usepackage{comment}
\usepackage{url,color}
\usepackage{paralist}

\newcommand\CL{\ensuremath{\mathcal{CL}}}
\newcommand\AnaCon{\ensuremath{\mathsf{AnaCon}}}

\title{From Contracts in Structured English to \CL\ Specifications}
\author{Seyed Morteza Montazeri, Nivir Kanti Singha Roy
\institute{Department of Computer Science and Engineering\\
University of Gothenburg, Sweden}
\email{gusmonse@student.gu.se, guskanthni@student.gu.se}
\and
Gerardo Schneider
\institute{Department of Computer Science and Engineering\\
Chalmers $\mid$ University of Gothenburg, Sweden\\
Department of Informatics\\
University of Oslo, Norway}
\email{gersch@chalmers.se}
}
\def\titlerunning{From Contracts in Structured English to \CL\ Specifications}
\def\authorrunning{Montazeri, Roy \& Schneider}

\begin{document}
\maketitle

\begin{abstract}
In this paper we present a framework to analyze conflicts of contracts written in structured English. A contract that has manually been rewritten in a structured English is automatically translated into  a formal language using the Grammatical Framework (GF). In particular we use the contract language \CL\ as a target formal language for this translation. In our framework \CL\ specifications could then be input into the tool CLAN to detect the presence of conflicts (whether there are contradictory obligations, permissions, and prohibitions. We also use GF to get a version in (restricted) English of \CL\ formulae. We discuss the implementation of such a framework.
\end{abstract}

\section{Introduction}  

Natural language (e.g., English) descriptions and prescriptions abound on documents used in different phases of the software development process, including informal specifications,  requirements, and contracts at different levels (methods/functions, objects, components, services,  etc.). There is no doubt of the usefulness of having such descriptions and prescriptions in natural language (NL), as most of the intended users of the corresponding documents would not have problems to understand them. However, it is well known that NL is ambiguous and imprecise in many cases due to context sensitivity, underspecified terminology, or simply bad use of the language.

At the other extreme, we have formal methods with a myriad of different formal languages (logics) with complex syntax and semantics. Those languages are indeed extremely useful as they are precise, unambiguous, and in many cases tools are available as to provide the possibility of (semi-) automatic analysis. However, in many cases they require high expertise not only at the syntactic level in order to use the language to specify system properties or requirements, but also to interpret the results of the tools. For instance, though the use of model checkers has been advertised as a ``push-button technology'' not requiring user expertise, the reality is that one still needs to write the properties on a logic and interpret the counter-examples which are also usually given as a big formula representing the trace leading to the problematic case.

On an ideal world, software engineers (and the mortal non-technical users) would only need to deal with natural language descriptions, push a button and get a result telling them whether for instance the given contract (specifications, set of requirements)\footnote{In the rest of the paper we will use the term ``contract'' to refer to contracts at different levels (including legal contracts), software specifications, requirements, etc.} is consistent and conflict-free. The current state-of-practice however is far from that ideal world. Though the state of the art on NL processing has advanced quite a lot in recent years, we still have to depend on the use of formal languages and techniques to analyze such contracts. A relatively new trend is then to restrict the use of NL in order to get something that ``looks like a NL'' but it has a better structure, and if possible avoid ambiguity. We call such constrained languages {\it restricted} or {\it controlled} NL.

Our aim in this paper is to advance towards finding a suitable solution to the problems and challenges just mentioned, in particular the possibility of writing contracts in NL, but being able to analyze them with tools, automating the process as much of possible.
In particular, we show that it is possible to relate the formal language for contracts \CL\  \cite{prisacariu9cl} and a restricted NL by using the Grammatical Framework (GF) \cite{GF}. In this way we are able to take contracts written in NL, manually obtain a restricted NL version of such contract, use GF to automatically obtain a \CL\
 formula that could then be analyzed using CLAN \cite{fenech2009clan}.
Note that the above process should be in both directions, as we might need the translation from 
\CL\  into NL since the result of CLAN in case a conflict is detected is given as a (eventually huge) \CL\ formula representing the counter-example. 

The paper is organized as follows. In next section we recall the necessary technical background the rest of the paper is based on, including \CL\ and GF. In section \ref{sec:frame} we present our framework in general terms, and we provide some details on the implementation. Before concluding in the last section we present a case study in section \ref{sec:case}.  

\section{Background}
\subsection{The Contract Language \CL}
                                                         
\CL\ is a logic based on combination of deontic, dynamic and temporal logics, designed to specify and reason about legal and electronic (software) contracts \cite{Prisacariu2007fle,prisacariu9cl}. With the help of \CL\ it is possible to  represent the deontic notions of obligation, permission and prohibition, as well as the penalties applied in case of not respecting the obligations and prohibitions. In what follows we recall the syntax of \CL, and we give a brief intuitive explanation of its notations and terminology, following \cite{prisacariu9cl}.  A contract in \CL\ may be obtained by using the syntax shown in Fig.~\ref{fig:cl}.


\begin{figure}[t]
\begin{tabular}{cc}
\begin{minipage}[c]{8cm}
\begin{eqnarray*}
 C &:= &C_O \mid C_P|C_F \mid C \wedge C \mid [\beta]C \mid \top \mid \bot \\
 C_O &:= &O_C(\alpha) \mid C_O \oplus C_O \\
 C_P &:= &P(\alpha) \mid C_P \oplus C_P
 \end{eqnarray*}
 \end{minipage} &
\begin{minipage}[c]{6.5cm}
\begin{eqnarray*}
 C_F &:= &F_C(\alpha) \\
 \alpha &:= &0 \mid 1 \mid a \mid \alpha \& \alpha|\alpha . \alpha \mid \alpha + \alpha  \\
 \beta &:= &0 \mid 1 \mid a \mid \beta \& \beta|\beta . \beta \mid \beta + \beta \mid \beta^*
\end{eqnarray*}
\end{minipage}
\end{tabular}
\caption{\CL\ syntax.\label{fig:cl}}
\end{figure}

A contract clause in \CL\ is usually defined by a formula $\mathit{C}$, which can be either an obligation ($\mathit{C_O}$), a permission ($\mathit{C_P}$) or a prohibition ($\mathit{C_F}$) clause, a conjunction of two clauses or a clause preceded by the dynamic logic square brackets. $\mathit{O}$, $\mathit{P}$ and $\mathit{F}$ are deontic modalities, the obligation to perform an action $\alpha$ is written as $\mathit{O_C(\alpha)}$, illustrating the primary obligation to perform $\alpha$, and if $\alpha$ is not performed then the reparation contract $\mathit{C}$ is enacted. The above represents in fact a {\it CTD} (\textit{Contrary-to-Duty}) as it specifies what is to be done if the primary obligation is not fulfilled. The prohibition to perform $\alpha$ is represented by the formula $\mathit{F_C(\alpha)}$, which not only specifies what is forbidden but also what is to be done in case the prohibition is violated (the contract $C$); this is called {\it CTP} ( \textit{Contrary-to-Prohibition}). Both CTDs and CTPs are then useful to represent normal (expected) behavior as well as the alternative (exceptional) behavior. $\mathit{P(\alpha)}$ represents the permission of performing a given action $\alpha$. 
In the description of the syntax, we have also represented what are the allowed actions ($\alpha$ and $\beta$ in Fig.~\ref{fig:cl}). It should be noticed that the usage of the Kleene star (the * operator) which is used to model repetition of actions, is not allowed inside the above described deontic modalities, though they can be used in dynamic logic style-conditions. Indeed, actions $\beta$ may be used inside the dynamic logic modality (the bracket $[\cdot]$) representing a condition in which the contract $\mathit{C}$ must be executed if action $\beta$ is performed. 
The binary constructors  \&, ., and + represent concurrency, sequence and choice in basic actions (e.g. ``buy'', ``sell'') respectively. Compound actions are formed from basic actions by using the above operators. Conjunction of clauses can be expressed using the $\wedge$ operator; the exclusive choice operator ($\oplus$) can only be used in a restricted manner. $\top$ and $\bot$ are the trivially satisfied and violating contract respectively. $0$ and $1$ are two special actions that represent the impossible action and the skip action (matching any action) respectively.\footnote{Some of the reasons behind the \CL\ syntax have been motivated by a desire to avoid deontic paradoxes \cite{McNamara2006}. See \cite{Prisacariu2007fle} for a more detailed explanation of the design decisions behind \CL.} 

The following example is an excerpt from part of a contract between an Internet provider and a client, where the provider gives access to the Internet to the client: ``The Client shall not supply false information to the Clients Relations Department of the provider'', and ``if the client does provide false information, the provider may suspend the service'' \cite{pace2007model}. If we consider the action
 $\mathit{fi}$ (representing that ``client supplies false information to Client Relations Department"), and  $s$ (representing that ``provider suspends service"), in \CL\ the above would be written as $F_{P(s)}(fi)$.

One of the usefulness of \CL\ is that there are tools available to detect whether a given \CL\ formula contains {\it conflicts}. 
There are four main kinds of conflicts in normative systems, and contracts in particular. The first one is when there is an obligation and the prohibition of performing the same action, in which case, independently of what  the performed action is, will lead to a violation of the contract. The second conflict type happens when there is a permission and a prohibition to do the same action, which might lead to a contradicting situation.
 The other two cases are when there is an obligation to perform mutually exclusive actions, and the permission and the obligation to perform mutually exclusive actions.
CLAN is a tool that automatically determines whether there are conflicts on \CL\ formulae, giving a counter-example in case a conflict is detected \cite{fenech2009clan}.

\subsection{Grammatical Framework}

The Grammatical Framework (GF) is a framework to define and manipulate grammars  \cite{GF}. Historically, GF was first implemented in the project ``Multilingual Document Authoring'' at Xerox Research Center Europe in Grenoble in 1998 \cite{GF}. As explained in \cite{hahnle2002authoring} the main idea of this project was to build an editor which helps a user to write a document in a language which is unfamiliar, as Italian,  while at the same time seeing how it develops in a familiar language such as English. The development of GF continued over time and evolved into a functional programming language, with multiple application domains \cite{GF}.
GF has a central data structure called {\it abstract syntax}, based on type theory where it is possible to define special purpose grammars. It also has a {\it concrete syntax} part where it is possible to specify how the formulas defined in the abstract syntax can be translated into NL or a formal notation. 
One notable use of GF useful for our purposes, is the possibility to relate NL and formal languages in both directions: it is possible to go from a sentence written in NL to a term in the formal language, and vice-versa.
GF has a module system consisting basically of two main modules, one to define the abstract syntax and the other the concrete syntax.       

\vspace*{1ex}
\noindent {\it Abstract syntax.}
Abstract syntax is a type-theoretical part of GF where logical calculi as well as mathematical theories may be defined simply by using type signatures \cite{hahnle2002authoring}. Let us consider the simple ``Hello world'' example. We start by defining the types of meaning (categories) \verb+Greeting+ and \verb+Recipient+: \verb+cat Greeting ; Recipient+.

\noindent We then define \verb Hello \space as a function for building syntactic trees. The type of \verb Hello \space has \verb Greeting \space as its value type and \verb Recipient \space as its argument type: \verb+fun Hello: Recipient -> Greeting;+.

\noindent {\it Concrete syntax.}
Once an abstract syntax is constructed, we can build a concrete syntax by using {\em linearization} rules, which basically allow us to generate the language. For instance, the English linearization rules for the function \verb+Hello+ is: \verb-lin Hello recip = {s = ``hello'' ++ recip.s};-.

\noindent The above defines the linearization of \verb Hello \space in terms of the linearization of its argument, represented by the variable \verb+recip+.
The terminal \verb ``hello'' \space is concatenated with \verb recip.s \space using the concatenation operator ++, which combines terminals. The most important thing to consider in a linearization rule is to define a string as a function of the variable it depends on. 

The example above shows that the \verb fun \space rules should be defined in the abstract syntax modules and \verb lin \space rules in concrete syntax modules. In the abstract syntax is where we need to define powerful type theory to express dependencies among parts of texts, and in the concrete syntax we need to define language-dependent parameter systems and the grammatical structures.

From the practical point of view, the concrete syntax description above could be saved on a file (let say \verb+example.gf+), which could be uploaded by making \verb+import example.gf+. By executing the command line \verb+Parse "hello world"+ we generate the abstract syntax \verb+Hello World+. It should be noticed that \verb+Hello+ is the function defined in the abstract syntax above (prefixed with \verb+fun+), and \verb+Recipient+ is the argument type of the function (i.e., \verb+World+ is of type \verb+Recipient+). In the concrete syntax, \verb+Recipient+ is a record \verb+recip+ containing a string \verb+s+ which in this case will take the value \verb+"world"+.

To summarize, the two main functionalities in GF useful for our purpose are {\it linearization} (translation from abstract to concrete syntax) and {\it parsing} (translation from concrete to abstract syntax). 

\section{Our framework} \label{sec:frame}

In this section we present our framework, \AnaCon, in general terms, and we present a summary of the linearization and parsing process of \CL\ into GF.


%
\subsection{The framework in general terms}

\AnaCon\ takes as input a text file containing the description of a contract consisting of 3 parts:
\begin{inparaenum}[1)]
\item A {\it Dictionary} listing the actions being used in the contract together with a textual description;
\item The contract itself written in restricted English;
\item A list of contradictory actions.
\end{inparaenum}
(Fig.~\ref{fig:input} shows the input file containing a simple contract.)

\begin{figure}
  \centering
  \includegraphics[scale=0.55]{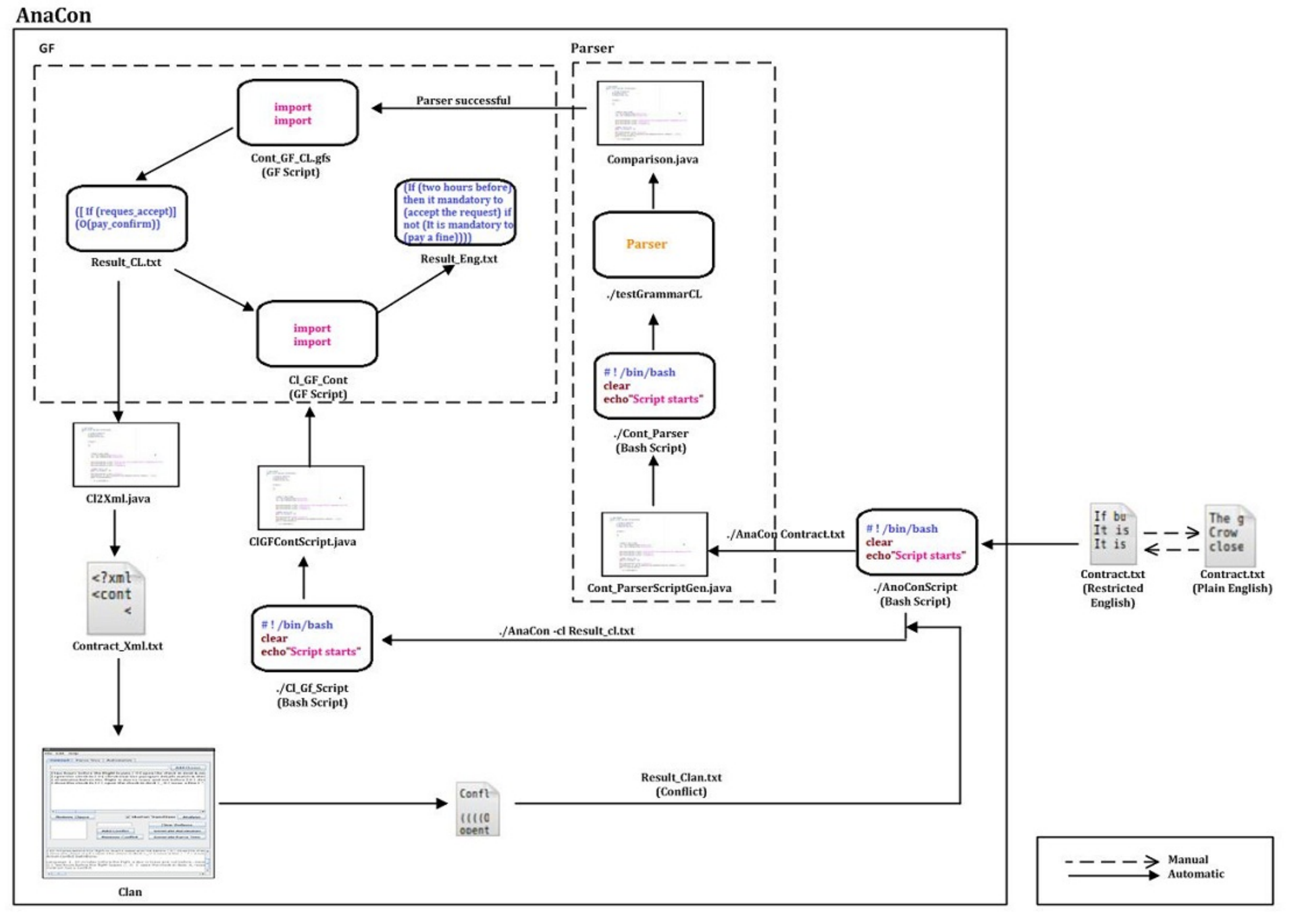}
  \caption{\AnaCon\ Framework}
   \label{fig:result}
\end{figure}

Our framework is summarized in Fig.~\ref{fig:result} where arrows represent the flow of information highlighting how it works. Essentially, it consists of a parser, the Grammatical Framework, the conflict analysis tool CLAN, and some scripts used to connect these different parts. Overall, the typical workflow of \AnaCon\ is as follows:

\begin{figure}
  \centering
  \includegraphics[scale=0.50]{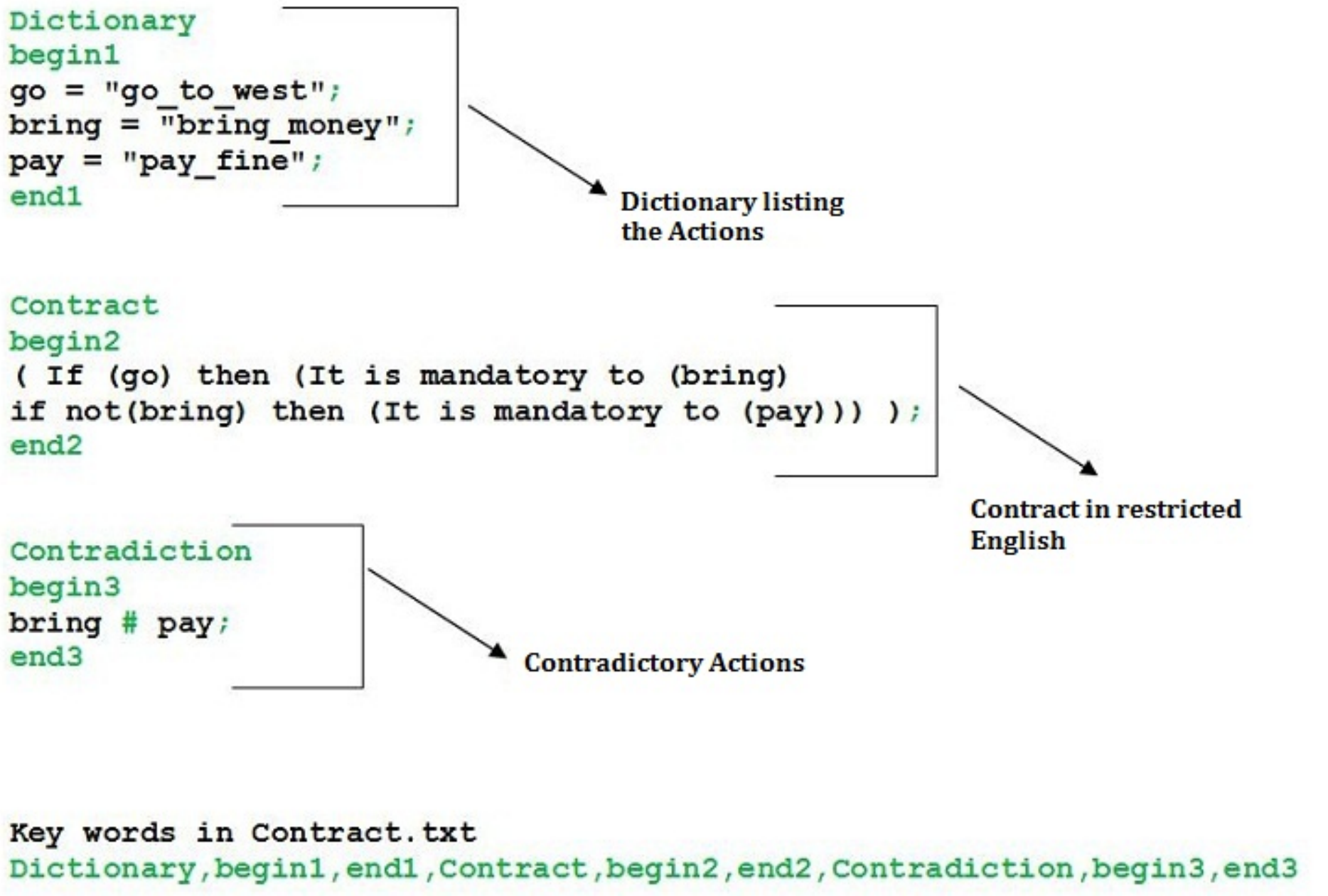}
  \caption{Sample Contract.txt}
   \label{fig:input}
\end{figure}

\begin{compactenum}
\item The user writes a contract (specification, set of requirements, etc) in NL (``plain'' English), which is then manually translated into restricted English. This is a modeling task and it is done manually. It does not require any technical skill from the user, only to get access to the list of ``allowed'' English words to be used in the restricted version of the language. For instance, a sentence originally written as ``The ground crew is obliged to open the check-in desk'' would be translated into ``It is mandatory to open the check-in desk'', where ``open the check-in desk'' is an action name representing the real action.
\item \label{AnaCon_script} The version of the contract written in restricted English is then passed to \AnaCon\ script as an argument so the analysis starts (in what follows we call the input file containing the given contract, {\it Contract.txt}).
\item \label{Cont_ParserScriptGen_java} \verb+Cont_ParserScriptGen+ (a Java program) generates a script file based on the content of {\it Contract.txt}. The script file \verb+Cont_Parser+ then projects the content of the file to \verb+testGrammarCl+ parser.
\item \label{testGrammarCl} At this stage \verb+testGrammarCl+ conducts syntax analysis based on the structure defined for the system. This parser is based on Labelled BNF grammar and generated from BNF Converter \cite{lbnf}.
\item \label{Comparison} The Java program \verb+Comparison+ connects with \verb+testGrammarCl+ in order to obtain actions defined in the contract and then compare these actions against the ones defined in the {\it Dictionary} part of  {\it Contract.txt}. Other analysis such as comparison between actions defined in {\it Contradiction} and the ones in {\it Dictionary}, duplication of actions in {\it Dictionary} and empty string assertion, are conducted at this level.
\item \label{Cont_GF_Cl.gfs} After successful parsing, the \verb+Cont_GF_Cl+ script file is generated with the contract and necessary information to start the translation process in GF from Restricted English to \CL.             
\item \label{RE_GF} The version of the contract written in restricted English is then represented in the abstract syntax part of GF.
\item \label{GF_CL} The abstract syntax obtained above is translated into concrete syntax (\CL) which is then stored in the text file {\it Result\_Cl.txt}.
\item The concrete syntax (\CL\ formula) in textual form is transformed into XML by using \verb+Cl2XML+ (implemented but not integrated).
\item The XML version of the \CL\ output of GF is fed into CLAN for analyzing whether the contract has a conflict.
\item If the output of CLAN is `NO', then the answer is immediately given to the user. If the answer is 'YES' then the counter-example will be given by CLAN (a big \CL\ formula containing the conflicting subclauses as well as the trace leading to such a state).

\item \label{CL_GF} The  formula obtained from CLAN is then linearized into a restricted English using GF. 
The \verb+Cl_GF+ script will take the content of {\it Result\_Cl.txt}  and pass it to \verb+Cl_GF_ContScript+ (a Java program), which generates the \verb+Cl_GF_Cont+ script to start the translation. The result in restricted English is given in a file named {\it Result\_Eng.txt}. 
\item The user must then find in the original contract where the counter-example arises. This last step is currently done manually, by simply searching in the text the keywords given in the counter-example in restricted English. (We discuss in the last section our future work on how to automate part of this process.)
\end{compactenum}

Except for the steps above where we explicitly mention that it is manual, the rest of the process is completely automatic. So far, we have only fully implemented steps \ref{AnaCon_script}--\ref{GF_CL}, and the translation back from \CL\ to restricted English (as used in step \ref{CL_GF}).

Besides the above, we provide the possibility of generating a restricted English version of a \CL\  formula, by executing \AnaCon\ with a special flag (\verb+AnaCon -cl <input_file.txt>+).

In what follows we will present some details of the implementation concerning the use of GF only. 

\subsection{Linearization and Parsing}

In what follows we present the abstract and concrete syntax of \CL\ and NL, in GF. At first, we present all the categories and functions for handling different \CL\ clauses and actions in the abstract syntax part, and then we concentrate on the concrete syntax part showing their representation in natural language. Due to lack of space we only show here some parts of the process.

The abstract syntax module as a central structure contains the basis for representation and formalization of \CL\ syntax. The linearization of the different used functions and structures into \CL\ symbolic syntax and natural language is done by using two concrete syntax modules. In the first one we write the exact \CL\ syntax, and in the latter we express the corresponding restricted natural language. 

We define the following categories, based on the BNF of  \CL.

{\small
\begin{verbatim}
 --  Abstract module(Cl.gf module)
cat  	 Act ;KleeneStarAct;KleeneCompAct;ClSO;ClSF;ClSP;Clause;Clauses;ClauseP;
	ClauseO; ClauseF;And;Or;Dot;Cross;CompAct;Star;Not;[Clause]{2};[CompAct]{2};
	[ClauseO]{2}; [ClauseP]{2};
\end{verbatim}
}
We define similar categories in the concrete modules \verb+ClEng.gf+ and \verb+ClSym.gf+, using \verb+lincat+ instead of \verb+cat+ prefixing them. The main difference is that in the linearization type definitions we need to  state that for instance \verb+Clause+ and \verb+Act+ are records containing a field of type string \verb+s+. 

\paragraph{Obligations, Permissions and Prohibitions.}
Obligations, permissions and prohibitions have the same structure in the \verb+Cl.gf+ module. In the rest of this section we will only show the abstract and concrete syntax for obligations, those for permissions and prohibitions being similar.
      
{\small
\begin{verbatim}
 --  Abstract module(Cl.gf module)
fun Obl : ClSO -> CompAct -> Clauses;       
    OClause : ClauseO -> Clause;;
    Clo : ClSO -> CompAct -> ClauseO;
\end{verbatim}
}
There are some differences among these clauses (or rather group of clauses, as there are similar ones for permissions and prohibitions) which is worth mentioning. As it is clearly shown \verb CompAct \space is the argument type used among the four groups which represent both basic and compound actions respectively. This will provide the possibility to be able to express \verb+Obligation+ over both basic and compound actions (actions will be explained later on in this section). \verb+ClauseO+ follows the BNF of the syntactic definition of \CL\ obligations and allows to express the obligation that together with the action (verb) form the actual clause.
The other difference is that generally a \verb Clause \space itself can consist of different structures and thus it can be either constructed from \verb+ClauseO+, \verb+ClauseP+ and \verb+ClauseF+ from basic or compound actions. They are defined in this way to avoid mixing certain operators only permitted for some of the deontic notions. It facilitates to make the conjunction and choice of certain kind of clauses, but not the direct linearization and parsing of each structure. For that we specify \verb+Clauses+ as a start category for parsing and linearization so that each structure can be linearized and parsed directly. The linearization of the above in concrete syntax is as follows:

{\small
\begin{verbatim}
 --  Concrete module(ClEng.gf module)
lin  Clo clo compact  = {s = "(" ++ clo.s ++ "(" ++ compact.s ++ ")" ++ ")"};
     OClause clo = {s = clo.s};
     Obl clo compact  = {s = "(" ++ clo.s ++ "(" ++ compact.s ++ ")" ++ ")"};  
\end{verbatim}
}
Basically, \verb+Clo+ and \verb+Obl+  clauses are expressed in NL using restricted words as ``It is mandatory to'' (\verb clo.s ) as terminals (quoted words in GF are {\it terminals}) so that together with actions they formulate the clauses in NL.

As explained before, we provide another concrete syntax module called \verb+ClSym.gf+ to provide the user with the possibility of writing with specific \CL\ syntax, such as operators, parentheses, brackets, etc. The converse is also true, when writing any specific clause in NL, the framework (with the help of this module) would be able to provide \CL\ formulas in the intended format.

{\small
\begin{verbatim}
 --  Concrete module(ClSym.gf module)
lin  Clo clo compact  = {s = "(" ++ clo.s ++ "(" ++ compact.s ++ ")" ++ ")"};
     OClause clo = {s = clo.s};
     Obl clo compact  = {s = "(" ++ clo.s ++ "(" ++ compact.s ++ ")" ++ ")"};
\end{verbatim}
}
The structure of the above module is very similar to the \verb+ClEng.gf+ module, the only difference being that \verb+clo.s+ represent specific characters such as \verb+"O"+ instead of words or sentences.

{\small
\begin{verbatim}
 --  Concrete module(ClSym.gf module)
lin O = {s = "O"};
\end{verbatim}
}

\paragraph{Contrary-to-Duties (CTDs) and Contrary-to-Prohibitions (CTPs).}  CTDs and CTPs are both related to obligation and prohibition clauses respectively, and are expressed as the following functions:

{\small
\begin{verbatim}
 --  Abstract module(Cl.gf module) 
fun  CTDc :  CompAct -> Clause -> Clauses;
     CTDcc: CompAct -> Clause -> Clause;
\end{verbatim}
}
CTD and CTP clauses are functions taking an action (which is to be obliged, or prohibited) and a clause representing what is to be done in case the obligation or the prohibition is not fulfilled. We specify again these operators over simple and compound actions. We only present now the concrete modules for CTDs, the ones for CTPs being similar.

{\small
\begin{verbatim}
 --  Concrete module(ClEng.gf module)
lin CTDc compact clause = {s = "(" ++ "It is mandatory to" ++ "(" ++ compact.s ++ ")" 
     ++"if not" ++ "(" ++ compact.s ++ ")"++ "then" ++ clause.s ++ ")"};
           
    CTDcc compact clause = {s = "(" ++ "It is mandatory to" ++ "(" ++ compact.s ++ ")"
      ++ "if not" ++ "(" ++ compact.s ++ ")" ++ "then" ++ clause.s ++ ")"};
	          
\end{verbatim}
}
Now, we show how to express the logical syntax in the other concrete module:

{\small
\begin{verbatim}
 --  Concrete module(ClSym.gf module)
lin CTDc compact clause = {s = "(" ++ "O" ++ "(" ++ compact.s ++ ")" ++ "_" 
     ++ clause.s ++ ")"};
    CTDcc compact clause = {s = "(" ++ "O" ++ "(" ++ compact.s ++ ")" ++ "_" 
    ++ clause.s ++ ")"};
	     
\end{verbatim}  
}
The only important thing to notice here is the \verb ``_'' \space character to express the reparation, meaning that the clause after this symbol is the reparation clause which has to be considered in case of a violation of the primary obligation.

\paragraph{Conjunction of clauses ($\mathit{C \wedge C}$)}
Other operators, like conjunction and exclusive or, need also to be represented in the abstract and concrete syntax. We only show here how to represent the conjunction.
   
{\small
\begin{verbatim}
 --  Abstract module(Cl.gf module)
fun Conj_np : [Clause] -> Clauses;       
    Conj_np2 : [Clause] -> Clause;
\end{verbatim}
}
As it is clearly defined in the above representation the structure used for conjunction of clauses consists of list of clauses which may be any kind of clause such as obligation, prohibition, etc. In this it is possible to define conjunction of many clauses. The two concrete modules are as follows:

{\small
\begin{verbatim}
 --  Concrete module(ClEng.gf module)
lin Conj_np xs = {s = "(" ++ xs.s ! Conjunction_np ++ ")"};
    Conj_np2 xs = {s = "(" ++ xs.s ! Conjunction_np ++ ")"};
\end{verbatim}
}
{\small
\begin{verbatim}
 --  Concrete module(ClSym.gf module)
lin Conj_np xs = {s = "(" ++ xs.s ! Conjunction_np ++ ")"};
    Conj_np2 xs = {s = "(" ++ xs.s ! Conjunction_np ++ ")"};
\end{verbatim}
}
The structure used in above to show iteration conjunction of clauses, corresponds to the way it was defined in \CL.

\paragraph{Test Operator.}
The test operator where in \CL\ is used to  express conditional obligations, permissions and prohibitions. The test operator may be applied to simple or compound actions including the Kleene star. We should thus add a function to each different application of the test operator; we will only present the abstract and concrete syntax of the application of the Kleene star to simple and compound actions. 

{\small
\begin{verbatim}
 --  abstract module(Cl.gf module)
fun TestOpc,TestOpcStar : KleeneCompAct -> Clause -> Clauses;         
    TestOpcc,TestOpccStar : KleeneCompAct -> Clause -> Clause;
\end{verbatim}    
}

The concrete modules are as follows:

{\small
\begin{verbatim}
 --  concrete module(ClEng.gf module)
  fun TestOpc kleenecompact clause = {s = "(" ++ "If" ++ "(" ++ kleenecompact.s ++")"
    ++ "then"++ clause.s ++ ")"} ;
                
  TestOpcStar kleenecompact clause = {s ="("++"((" ++ "Always" ++ ")"|"(" ++ "After" 
    ++ ")"|"(" ++ "When" ++ ")"|"(" ++ "Before" ++ ")") ++ "(" ++ "If" ++ "(" ++
    kleenecompact.s ++ ")" ++ "then"++ clause.s ++ ")" ++ ")"} ;
              
  TestOpcc kleenecompact clause = {s = "(" ++ "If" ++ "(" ++ kleenecompact.s ++ ")" 
    ++ "then"++ clause.s ++ ")"};

   TestOpccStar kleenecompact clause = {s = "(" ++ ( "(" ++ "Always" ++ ")"|"(" ++ 
    "After" ++ ")"|"(" ++ "When" ++ ")"| "(" ++ "Before" ++ ")") ++ "(" ++ "If" ++ 
    "(" ++ kleenecompact.s ++ ")" ++  "then"++ clause.s ++ ")" ++ ")"};
\end{verbatim}  
}

{\small
\begin{verbatim}
 --  concrete module(ClSym.gf module)
   fun TestOpc kleenecompact clause = {s = "(" ++ "[" ++ "(" ++ kleenecompact.s ++")" 
     ++ "]" ++ clause.s ++ ")"} ;
    
   TestOpcStar kleenecompact clause = {s = "(" ++ "[" ++ "1" ++ "*" ++ "]" ++
    "(" ++ "[" ++ "(" ++ kleenecompact.s ++ ")" ++ "]" ++ clause.s ++ ")" ++ ")"} ;
       
   TestOpcc kleenecompact clause = {s = "(" ++ "[" ++ "(" ++ kleenecompact.s
   ++ ")" ++ "]" ++ clause.s ++ ")"};
	      
   TestOpccStar kleenecompact clause = {s = "(" ++ "[" ++ "1" ++ "*" ++ "]"
    ++ "(" ++ "[" ++ "(" ++ kleenecompact.s ++ ")" ++ "]" ++ clause.s ++ ")"++ ")"};
\end{verbatim}  
}

\paragraph{Actions.}
Defining basic, compound and Kleene star actions in GF is not difficult, however it should be noted that since actions in our case are generally verbs that could be considered as a specific vocabulary part (domain lexicon), it is more efficient to use module extension \cite{GF}. This in effect separates the grammar part (\verb+Cl.gf+  module) from a more specific vocabulary part (\verb+Action+ module). In other words, the developer will be provided with a modular system giving more flexibility to modify the modules, and thus increasing maintainability. The \verb+Action+ module extends the \verb+Cl+ module. In such a module we will define all the involved simple actions (e.g., ``Pay), other actions only affected by the Kleene star (e.g., ``CloseCheckIn"), as well as those operators over actions. In what follows we only show the conjunction (sequence of actions, choice, etc are defined similarly).

{\small
\begin{verbatim}
 --  Action module(abstract syntax)
fun Pay,Buy : Act;
\end{verbatim}
}
{\small
\begin{verbatim}
 --  Cl module (abstract syntax)
fun CompActSI : Act -> CompAct; 
    CompActa : CompAct -> And -> CompAct -> CompAct;
\end{verbatim}
} 
The linearization of the functions specified in the \verb+Action+  module above described is easy to specify:

{\small
\begin{verbatim}
-- ActionEng module (concrete syntax)
lin Pay = {s = "pay a fine"};
    Buy = {s = "buy a car"};
    CloseCheckIn = {s = "closeTheCheckIn"};
    CorrectDetail = {s = "checkThatThePassportDetailMatch"};  
\end{verbatim}
}
The structure of compound actions shows how the operators' name has been used as an argument types to build the functions.   However, what we need to focus on in the translation of compound actions into NL is to know how each operator should be interpreted. As a consequence we end up with the following concrete syntax where it is possible to express all the operators:

{\small
\begin{verbatim}
-- Cl module (abstract syntax)
fun CompActSI : Act -> CompAct; 
    CompActa : CompAct -> And -> CompAct -> CompAct;
\end{verbatim}
}
{\small
\begin{verbatim}
-- ClEng module (concrete syntax)
lin CompActSI acti  = {s = acti.s}; 
    CompActa compact and compact1 = {s = compact.s ++ and.s ++ compact1.s};
\end{verbatim}
}
User defined operations such as the above fall under specific logical symbols which are defined below:

{\small
\begin{verbatim}
-- ClSym module (concrete syntax)
lin  CompActa CompAct and CompAct1 = {s = CompAct.s ++ and.s ++ CompAct1.s};
     CompActSI acti  = {s = acti.s};
\end{verbatim}
}
In this manner, the representation of \verb+and.s+ (similarly for \verb+or.s+ , \verb+dot.s+, and \verb+not.s+) are reduced to \verb+"&"+, \verb-"+"-, \verb+"."+ and \verb+"!"+ which are logical operators as used in CLAN to manipulate \CL\ formulae.

The Kleene star is actually a compound action as shown below with the difference that it can only be used between test operator: 

{\small
\begin{verbatim}
-- Cl module (abstract syntax)
fun  KleeneActSI : Act -> KleeneCompAct;
     KleeneActa : KleeneCompAct -> And -> KleeneCompAct -> KleeneCompAct;
\end{verbatim}
}    
In what follows we show the corresponding concrete syntax enabling the translation to natural and symbolic languages:

{\small
\begin{verbatim}
-- ClEng module (concrete syntax)
lin  KleeneActSI acti = {s = acti.s};
     KleeneActa kleenecompact and kleenecompact1  = {s = kleenecompact.s ++ 
	and.s ++ kleenecompact1.s};
\end{verbatim}
}
{\small
\begin{verbatim}
-- ClSym module (concrete syntax)
lin  KleeneActSI acti = {s = acti.s};
     KleeneActa kleenecompact and kleenecompact1  = {s = kleenecompact.s 
	++ and.s ++ kleenecompact1.s};
\end{verbatim}
}

\section{Case Study}\label{sec:case} 

In this section we apply our framework to a case study taken from  \cite{fenech2009automatic} of a contract concerning the check-in process of an airline company. The full description is given in Fig.~\ref{fig:checkin}.
We provide the detailed translation into restricted English, and the corresponding \CL\ formula for all the clauses. Note that clause 10 is ``distributed" among the others as it represent a penalty in case the other clauses are not satisfied.

\begin{figure}[t]
{\small
\begin{enumerate}
 \item \textit{The ground crew is obliged to open the check-in desk and request the passenger manifest 
two hours before the flight leaves.}
 \item \textit{The airline is obliged to reply to the passenger manifest request made by the ground crew
when opening the desk with the passenger manifest.}
 \item \textit{After the check-in desk is opened the check-in crew is obliged to initiate the check-in process
with any customer present by checking that the passport details match what is written on
the ticket and that the luggage is within the weight limits. Then they are obliged to issue the
boarding pass.}
 \item \textit{If the luggage weighs more than the limit, the crew is obliged to collect payment for the extra
weight and issue the boarding pass.}
 \item \textit{The ground crew is prohibited from issuing any boarding cards without inspecting that the
details are correct beforehand.}
 \item \textit{The ground crew is prohibited from issuing any boarding cards before opening the check-in
desk.}
 \item \textit{The ground crew is obliged to close the check-in desk 20 minutes before the flight is due to
leave and not before.}
 \item \textit{After closing check-in, the crew must send the luggage information to the airline.}
 \item \textit{Once the check-in desk is closed, the ground crew is prohibited from issuing any boarding
pass or from reopening the check-in desk.}
 \item \textit{If any of the above obligations and prohibitions are violated a fine is to be paid.}
\end{enumerate}
}
\caption{Case study\label{fig:checkin}}
\end{figure}

\begin{enumerate}
 \item \textrm{The ground crew is obliged to open the check-in desk and request the passenger manifest two hours before the flight leaves (Fig.~\ref{fig:checkin} first clause). }\newline\newline
  \textit{[Restricted English]:  ( If (two\_hours\_before\_the\_flight\_leaves) then (It is mandatory to (open\_the\_ check\_in\_desk\_and\_request\_the\_passenger\_manifest) if not (open\_the\_check\_in\_desk\_and\_request\_
  the\_passenger\_manifest) then (It is mandatory to (pay\_a\_fine))  ) )\newline\newline}
  [Program output]: ( [ ( two\_hours\_before\_the\_flight\_leaves ) ] ( O ( open\_the\_check\_in\_desk \& request\_the
  \_ passenger\_manifest ) \_ ( O ( pay\_a\_fine ) ) ) )\\
  
\item \textrm{The airline is obliged to reply to the passenger manifest request made by the ground crew when opening the desk with the passenger manifest}\newline\newline
  \textit{[Restricted English]: ( (When) ( If (opening\_the\_desk\_with\_the\_passenger\_manifest) then (It is mandat
  ory to (reply\_to\_the\_passenger\_manifest\_request) if not (reply\_to\_the\_passenger\_manifest\_
  request) then (It is mandatory to (pay\_a\_fine))) )) \newline\newline}
  [Program output]: ( [ 1 * ] ( [ ( opening\_the\_desk\_with\_the\_passenger\_manifest ) ] ( O ( reply\_to\_the\_ pass
  enger\_manifest\_request ) \_ ( O ( pay\_a\_fine ) ) ) ) )\\
  
\item \textrm{After the check-in desk is opened the check-in crew is obliged to initiate the check-in process with any customer present by checking that the passport details match what is written on the ticket and that the luggage is within the weight limits. Then they are obliged to issue the boarding pass (Fig.~\ref{fig:checkin} third clause).}\newline\newline
\textit{[Restricted English]: (((After) (If (open\_the\_check\_in\_desk) then (It is mandatory to (check\_that\_the
\_ passport\_details\_match\_ what\_is\_written\_on\_the\_ticket\_and\_check\_
the\_luggage\_is\_within\_the\_ \\
weight\_limits) if not (check\_that\_the\_passport\_details\_match\_what\_is\_written\_on\_the\_ticket\_and\_ \\
check\_the\_luggage\_is\_within\_the \_weight\_limits) then (It is mandatory to (pay))))) and (If (check\_ \\
that\_the\_passport\_details\_match\_what\_is\_written\_on\_the\_ticket\_and\_check\_the\_luggage\_is\_within \\
\_the\_weight\_limits) then (It is mandatory to (issue\_the\_boarding\_pass) if not (issue\_the\_boarding\_ \\
pass) then (It is mandatory to (pay\_a\_fine)) ) ) )\newline\newline}	
[Program output]: (( [ 1 * ] ([(open\_the\_check\_in\_desk ) ] (O (check\_that\_the\_passport\_details\_
match\_what\_is\_written\_on\_the\_ticket \& check\_the\_luggage\_is\_within\_the\_weight\_limits) \_ (O (pay\_ \\
a\_fine ))))) $\wedge$ ([( check\_that\_the\_ passport\_details\_match\_what\_is\_written\_on\_the\_ticket \& check\_ \\
the\_luggage\_is\_within\_the\_weight\_limits) ] (O (issue\_the\_boarding\_pass) \_ (O (pay\_a\_fine )))))\\

\item \textrm{If the luggage weighs more than the limit, the crew is obliged to collect payment for the extra weight and issue the boarding pass (Fig.~\ref{fig:checkin} seventh clause).}\newline\newline
 \textit {[Restricted English]: ( If (the\_luggage\_weighs\_more\_than\_the\_limit) then (It is mandatory to (colle
ct\_ payment\_for\_the\_extra\_weight\_and\_issue\_the\_boarding\_pass ) if not (collect\_payment\_for\_the\_ \\
extra\_weight\_ and\_issue\_the\_boarding\_pass ) then (It is mandatory to (pay\_a\_fine))) )\newline\newline}
[Program output]:( [ ( the\_luggage\_weighs\_more\_than\_the\_limit ) ] ( O ( collect\_payment\_for\_the\_ \\
extra\_weight \& issue\_the\_boarding\_pass ) \_ ( O ( pay\_a\_fine ) ) ) )\\

\item \textrm{The ground crew is prohibited from issuing any boarding cards without inspecting that the details are correct beforehand. (Fig.~\ref{fig:checkin} ninth clause).}\newline\newline
 \textit {[Restricted English]: (It is mandatory to (inspect\_that\_the\_details\_are\_correct\_beforehand) if not (inspect\_that\_the\_details\_are\_correct\_beforehand) then (It is prohibited to (issue\_any\_boarding\_ \\
 cards) if (issue\_any\_boarding\_cards) then (It is mandatory to (pay\_a\_fine))) )\newline\newline}
[Program output]:( O ( inspect\_that\_the\_details\_are\_correct\_beforehand ) \_ ( F ( issue\_the\_boarding\_ \\
pass ) \_ ( O ( pay\_a\_fine ) ) ) )\\

 \item \textrm{The ground crew is prohibited from issuing any boarding cards before opening the check-in desk (Fig.~\ref{fig:checkin} tenth clause).}\newline\newline
  \textit{[Restricted English]: ( (Before) (If (open\_the\_check\_in\_desk) then (It is prohibited to (issue\_any\_ boarding\_cards) if (issue\_any\_boarding\_cards) then (It is mandatory to (pay\_a\_fine))  )  )  )\newline\newline}
[Program output]: ( [ 1 * ] ( [ ( open\_the\_check\_in\_desk ) ] ( F ( issue\_the\_boarding\_pass ) \_ ( O ( pay\_a\_fine ) ) ) ) )\\
  
  \item \textrm{The ground crew is obliged to close the check-in desk 20 minutes before the flight is due to leave and not before}\newline\newline
   \textit{[Restricted English]:( (Before) (If (20\_minutes\_the\_flight\_is\_due\_to\_leave\_and\_not\_before)  then (It is mandatory to (close\_ the\_check\_in\_desk) if not (close\_the\_check\_in\_desk) then (It is mandatory to (pay\_ a\_fine))  )  )  )\newline\newline}
   [Program output]:( [ 1 * ] ( [ ( 20\_minutes\_the\_flight\_is\_due\_to\_leave\_and\_not\_before ) ] ( O ( close\_the\_check\_in desk ) \_ ( O ( pay\_a\_fine ) ) ) ) )\\    
   
   \item \textrm{After closing check-in, the crew must send the luggage information to the airline}\newline\newline
    \textit{[Restricted English]:( (After) (If (close\_the\_check\_in\_desk) then (It is mandatory to (send\_the\_ \\
    luggage\_information\_to\_airline) if not (send\_the\_luggage\_information\_to\_airline) then (It is \\
    mandatory to (pay\_a\_fine))  )  )  )\newline\newline} 
    [Program output]:( [ 1 * ] ( [ ( close\_the\_check\_in\_desk ) ] ( O ( send\_the\_luggage\_information\_to\_ \\
    airline ) \_ ( O ( pay\_a\_fine ) ) ) ) )\\
    
    \item\textrm{Once the check-in desk is closed, the ground crew is prohibited from issuing any boarding pass or from reopening the check-in desk}\newline\newline
     \textit{[Restricted English]:( (Always) (If (close\_the\_check\_in\_desk)  then (It is prohibited to (issue\_any\_ \\
     boarding\_pass\_or\_open\_the\_check\_in\_desk) if (issue\_any\_boarding\_pass\_or\_open\_the\_check\_in\_ \\
     desk) then (It is mandatory to (pay\_a\_fine))  )  )  )\newline\newline}
     [Program output]:( [ 1 * ] ( [ ( close\_the\_check\_in\_desk ) ] ( F ( issue\_the\_boarding\_pass + open\_the\_ \\
     check\_in\_desk ) \_ ( O ( pay\_a\_fine ) ) ) ) )

\end{enumerate}

The case above has already been analyzed using CLAN before and a conflict has been detected as reported in \cite{fenech2009clan}. So, in this sense we do not report any new result here. 
We have used the same example as our intention is to validate our approach on a familiar case study when all the steps in our framework be implemented (we are currently working on a full implementation of the framework).

\section{Related Work}
GF has been used on a variety of application domains. We will only focus here on the one reported in \cite{hahnle2002authoring} since it is closely related to our research. In such paper  H\"ahnle et al. describes how to get a NL version of a specifications written in OCL (Object Constraint Language). The paper focused on helping to solve problems related to authoring well-formed formal specifications, maintaining them, mapping different levels of formality and synchronizing them. The solution outlined in the paper illustrates the feasibility of connecting specification languages at different levels, in particular OCL and NL. The authors have implemented different concepts of OCL such as classes, objects, attributes, operations and queries in GF. Our work is similar to \cite{hahnle2002authoring} with the difference that \CL\  is a more abstract and general logic allowing to specify contracts in a general sense (as mentioned in the introduction \CL\  may formalize legal contracts, software specifications, contracts in SOA, or even be used to represent requirements). Besides we are not interested only on ``language translation'' but rather in the use of the formal language to further perform verification (in our case conflict analysis) which is then integrated within our framework by connecting GF's output into CLAN. In what concerns the technical difficulties related to the implementation in GF we do not have enough knowledge of the work done in \cite{hahnle2002authoring} as to make a more careful comparison.

From the perspective of relating a contract language and natural language, it is worth mentioning the work by Pace and Rosner \cite{pace-controlled}, where it is presented an end-user system which is specifically designed to process the domain of computer oriented contracts. The translation is not based on GF but on a completely different technology. They use controlled natural language (CNL) to specify contracts, and define a similar logic to \CL, which is embedded into  Haskel in order to manipulate the contracts. So, the comparison is not straightforward as the aim of their work and ours diverges. 

\section{Conclusions}
We have presented in this paper an encoding of the contract language \CL\ into GF, and back. We have integrated the above into a framework that allows to analyze \CL\ formulae for conflicts, and eventually give a counter-example in restricted English helping the user to find it in the original specification in natural language. As a proof-of-concept we have applied it to a case study which has already been used for conflict detection.
  The framework as presented here does not automate the whole process, though we are working on those parts as described below.

We would like to emphasize what was said in the introduction concerning the scope of our approach. \CL\ is a formal language to specify contracts in a broad sense, and as such one should not think that our work limits to the analysis of contracts in that language. As an abstract logic, \CL\  can be used to describe and prescribe ``contracts'' (including specifications) in SOA, component-based development systems, e-business, requirement engineering, etc. 
We believe the approach is useful in practice, the only potential bottleneck being CLAN since the current version is not optimized as to obtain small non-redundant automata (the tool is very much a specialized explicit model checker, where high number of transitions are generated due to the occurrence of concurrent actions). 

One practical way to reduce the size of the automaton created by CLAN is to try to define as many mutually exclusive actions as possible. 
Note that some of the actions in our contract example are obviously mutually exclusive (e.g., `open the check in desk' and `close the check in desk' ), while others are mutually exclusive in the ``formal'' sense, that is we know that they cannot occur at the same time (for instance, `issue a fine' and `issue the boarding pass').
We are currently working on more fundamental ways to improve the performance of CLAN by reducing the size of the automaton while building it. 

A challenging future work concerns the use of Passage Retrieval tools (as for instance the one presented in \cite{buscaldi2009answering}) to help to find the counter-example in the original English contract by using the information in restricted English (obtained from CLAN and translated into English by our framework). This will avoid to manually go through big part of the English text, increasing efficiency and precision.
Another interesting line of research is to study how to combine our approach to the one presented in  \cite{pace-controlled}. We believe we could then improve our analysis capability, by using specialized tools for some purposes (as we have done here, using CLAN), and the embedded language technology for others.



\end{document}